\pdfoutput=1

\documentclass[11pt]{article}

\usepackage[]{EMNLP2022}

\usepackage{times}
\usepackage{latexsym}
\usepackage{graphicx}
\usepackage{stfloats}
\usepackage{amsfonts}
\usepackage{amsmath}
\usepackage{multirow}
\usepackage{enumitem}
\usepackage{makecell}
\usepackage{float}  
\usepackage{subfigure}  
\usepackage{paralist,bbding,pifont}
\usepackage[T1]{fontenc}

\usepackage[utf8]{inputenc}

\usepackage{microtype}

\usepackage{inconsolata}

\title{IRRGN: An Implicit Relational Reasoning Graph Network for Multi-turn Response Selection}

\author{Jingcheng Deng\textsuperscript{\rm 1,$\dagger$,} Hengwei Dai\textsuperscript{\rm 1,$\dagger$}, Xuewei Guo\textsuperscript{\rm 2,$\dagger$}, Yuanchen Ju\textsuperscript{\rm 1}, Wei Peng\textsuperscript{\rm 3,4,\thanks{~~Corresponding author. $\dagger$Equal contribution.}}\\
\textsuperscript{\rm 1}College of Computer and Information Science, Southwest University\\
\textsuperscript{\rm 2}yz-intelligence Inc\\
\textsuperscript{\rm 3}Institute of Information Engineering, Chinese Academy of Sciences, Beijing, China\\
\textsuperscript{\rm 4}School of Cyber Security, University of Chinese Academy of Sciences, Beijing, China\\
{\tt djc123234@163.com, kirobrine2000@163.com, g909336740@gmail.com} \\
{\tt juyuanchen0213@163.com, pengwei@iie.ac.cn}
}

\begin{document}
\maketitle
\makeatletter
\newcommand{\rmnum}[1]{\romannumeral #1}
\newcommand{\Rmnum}[1]{\expandafter\@slowromancap\romannumeral #1@}
\makeatother
\begin{abstract}
The task of response selection in multi-turn dialogue is to find the best option from all candidates. In order to improve the reasoning ability of the model, previous studies pay more attention to using explicit algorithms to model the dependencies between utterances, which are deterministic, limited and inflexible. In addition, few studies consider differences between the options before and after reasoning. In this paper, we propose an Implicit Relational Reasoning Graph Network to address these issues, which consists of the Utterance Relational Reasoner (URR) and the Option Dual Comparator (ODC). URR aims to implicitly extract dependencies between utterances, as well as utterances and options, and make reasoning with relational graph convolutional networks. ODC focuses on perceiving the difference between the options through dual comparison, which can eliminate the interference of the noise options. Experimental results on two multi-turn dialogue reasoning benchmark datasets MuTual and MuTual$^{plus}$ show that our method significantly improves the baseline of four pre-trained language models and achieves state-of-the-art performance. The model surpasses human performance for the first time on the MuTual dataset. Our code is released in the link.\footnote{The codes are available at: \small{\url{https://github.com/DJC-GO-SOLO/IRRGN}}}
\end{abstract}

\section{Introduction}
The response selection task is one of the most important tasks in neural dialogue systems \citep{DBLP:conf/acl/WelleckWSC19,DBLP:conf/acl/DemszkyMKCNR20,DBLP:journals/corr/abs-2204-12749,DBLP:conf/coling/ZhangCXX20,chen2021gog,DBLP:conf/ijcai/00080XXSL22,DBLP:conf/cec/PengHXXS22,DBLP:conf/ijcai/ZhaoZL22}, which aims to find the most appropriate response from a set of candidate options given a historical dialogue. Most previous studies focus on matching between candidate options and historical dialogues while ignoring the reasoning ability of the model. This causes the model to choose logically incorrect or even counter-common-sense options, resulting in a poor user experience \citep{DBLP:journals/jzusc/ShumHL18}. On the recently released multi-turn dialogue reasoning benchmark dataset MuTual \citep{DBLP:conf/acl/CuiWLZZ20}, these traditional representative response selection models\citep{DBLP:conf/acl/WuWXZL17,DBLP:conf/acl/WuLCZDYZL18} perform poorly, which indicates that comparing with matching, reasoning ability is more important in MuTual. Specifically, matching finds semantically related candidates, while reasoning requires obtaining logically consistent responses based on logical and semantic dependencies between sentences.

\begin{figure}[t]
	\centering
	\includegraphics[scale=0.35]{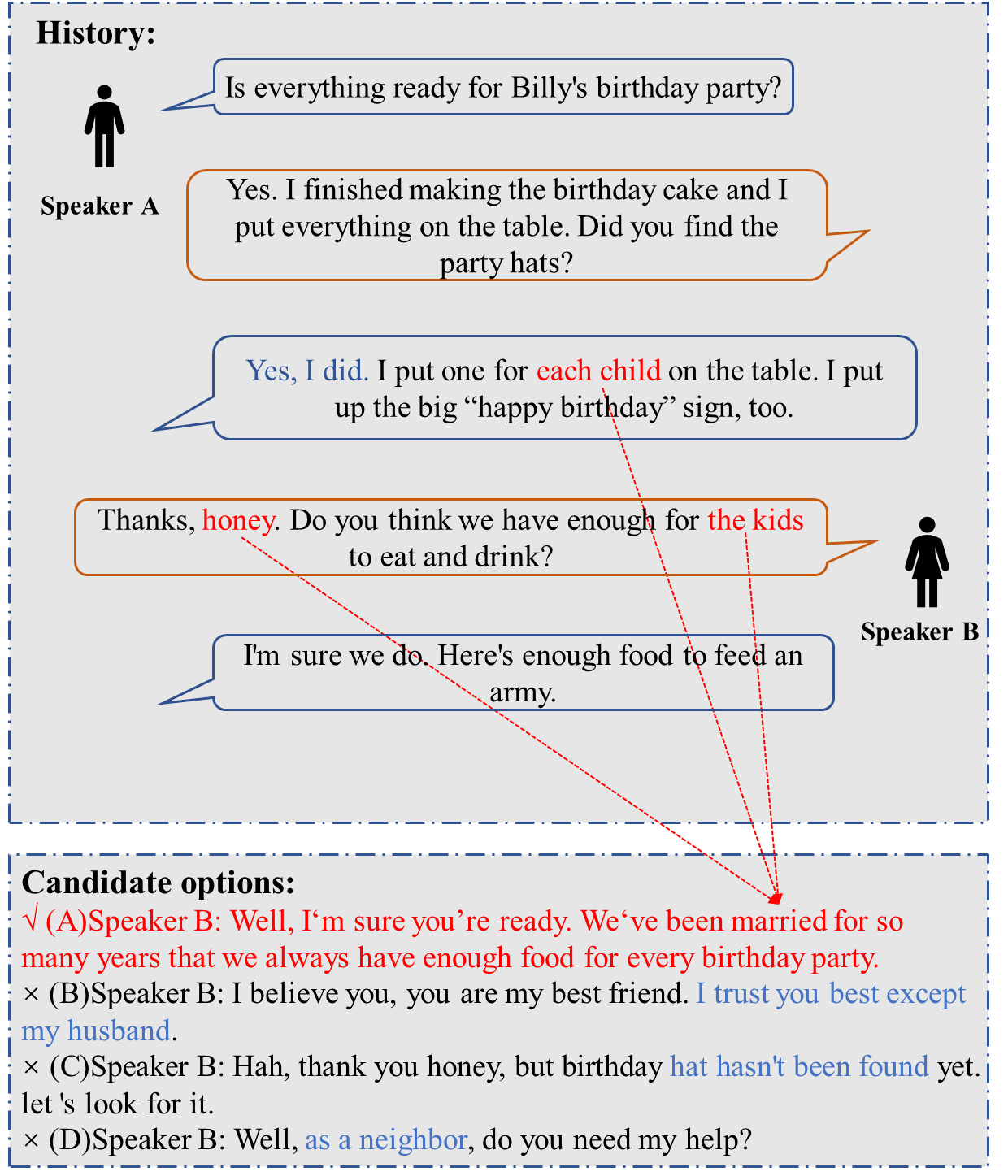}
	\caption{An example from MuTual. All candidate options are semantically related to the historical dialogue. Logical contradictions are marked with sky blue. Ground truth is marked with red. And the reasoning is red dashed lines.}
	\label{fig1}
\end{figure}
 For example, in Figure 1, all the relevant words that appear in option C include "\textit{honey}" and "\textit{birthday}", both of which occur in historical dialogue. Since traditional models tend to choose the more semantically relevant option, they consider option C as the best option. However, option C is not logically consistent with the historical dialogue. This conflicts with option C because the hat has already been found in the history dialogue. Similarly, options B and D also have the above problems. For the option A, "\textit{married}" can be inferred from "\textit{each child}", "\textit{honey}" and "\textit{the kids}", and it would be considered by a model with good reasoning ability. There are some work has emerged to improve the reasoning ability of models, but they still have some shortcomings. 

Firstly, modeling the dependencies between sentences is an important part of improving the reasoning ability of the model. Experience shows that temporal dependencies \citep{DBLP:conf/acl/LuZXLZX19,DBLP:conf/acl-mrqa/YehC19} between sentences as well as semantic dependencies are critical for multi-turn response selection. Previous studies typically  historical dialogues and candidate options as the context \citep{DBLP:conf/acl/SuSZSHNZ19}, or process each utterance independently \citep{DBLP:conf/acl/TaoWXHZY19}, which lead to ignoring dependencies between sentences. There are also methods to model dependencies through explicit rules, such as using some sentiment dictionaries to obtain features or relying on some community detection algorithm for modeling \citep{DBLP:conf/aaai/LiuFWSR021}. However, the dependencies established by these explicit rules are deterministic and limited, which affects the performance of the model and can not establish the flexible dependencies between the utterances.

In addition, previous models only focus on the reasoning method, while ignoring the difference between candidate options before and after reasoning. Generally, people first compare the candidate options to understand each of them. After that, people read the article or historical dialogue information to make a deep reasoning and finally make a correct comparison between the candidate options again. Taking Figure 1 as an example, humans first compare the differences among the four options and find that the relationship between speakers differs the most. Then based on historical conversation information, words such as "\textit{child}", "\textit{kids}" and "\textit{honey}" are captured. Comparing the differences between the four options again, it is found that only option A matches the historical dialogue, and finally, conclude that A is the best option. Inspired by human behaviors in reasoning, one can easily come to the correct answer after a two-stage comparison, which is similar to the preview and read methods in the PQ4R learning strategy.

Based on the above ideas, in this paper, we propose Implicit Relational Reasoning Graph Network (IRRGN), which consists of the Utterance Relation Reasoner (URR) and the Option Dual Comparator (ODC). Specifically, the purpose of the URR is to reason and adaptively capture flexible dependencies between utterances, as well as utterances and options, without relying on any explicit algorithm. The ODC is used to perceive the difference between the options before and after reasoning, which can eliminate the interference of the noise options. In summary, our contributions are as follows:


\begin{compactitem}
	\item We propose an URR, which adaptively captures flexible dependencies between utterances, as well as utterances and options, through a relational attention mechanism, and enables reasoning by propagating messages along various utterance paths.
	\item We propose an ODC, which captures the difference between options before and after reasoning according to the way humans think, which can eliminate the interference of the noise options.
	\item Empirical results show that our proposed model achieves state-of-the-art performance on MuTual and Mutual$^{plus}$ datasets. This is the first time the model surpass human performance on the MuTual dataset.
\end{compactitem}
\begin{figure*}[t]
	\centering
	\includegraphics[scale=0.38]{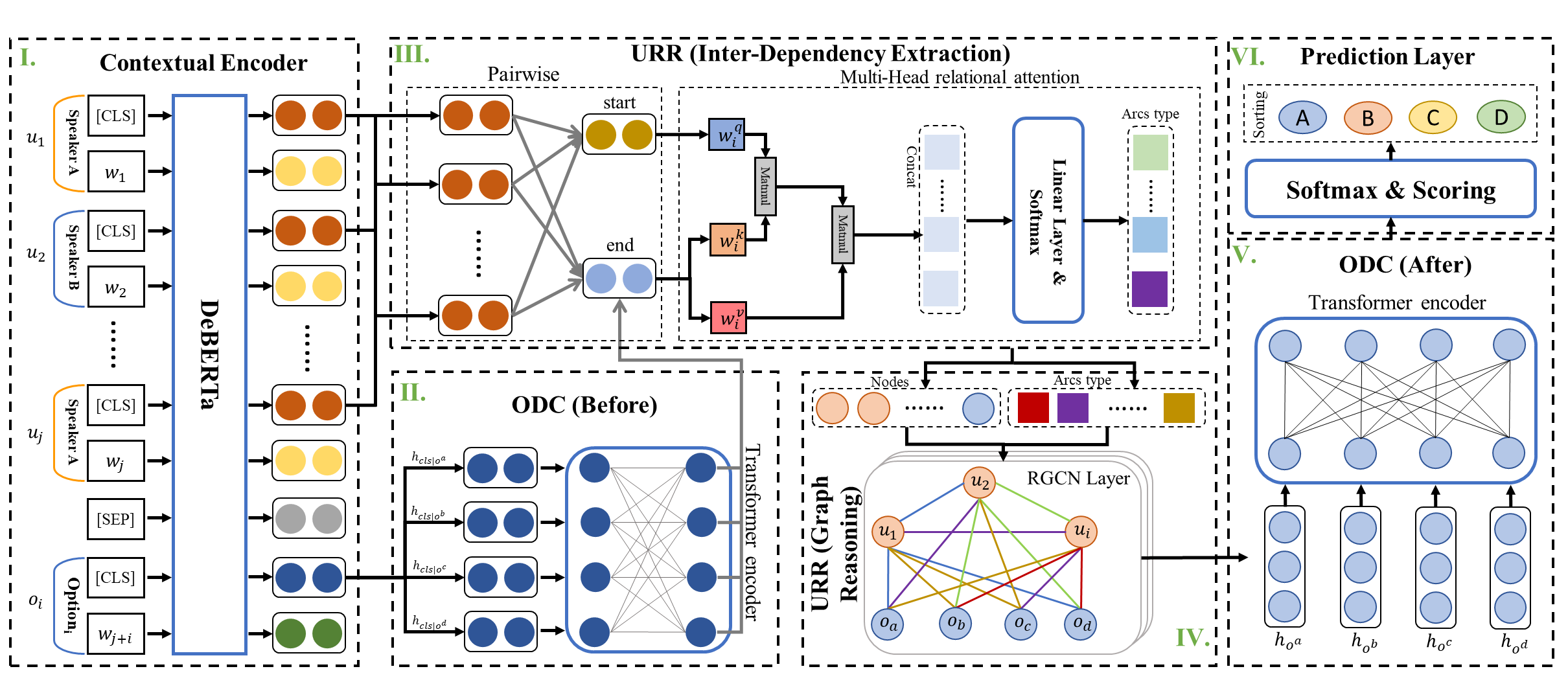}
	\caption{Overview of the proposed IRRGN. It contains six components: \Rmnum{1}. Contextual Encoder, \Rmnum{2}. ODC (Before), \Rmnum{3}. URR (Inter-Sentence Dependency Extraction), \Rmnum{4}. URR (Graph Reasoning), \Rmnum{5}. ODC (After), and \Rmnum{6}. Prediction. The gray arrow in \Rmnum{2} and \Rmnum{3} indicates that one of the elements is selected for calculation.}
	\label{fig2}
\end{figure*}
\section{Related Work}
\subsection{Response Selection Model}
Current research can be roughly divided into three categories \citep{DBLP:conf/ijcai/TaoF0WJ21}, which are representation-based models, interaction-based models, and pre-trained language model (PLM)-based models \citep{DBLP:conf/naacl/DevlinCLT19}. Representation-based models usually first encode historical conversations and candidate options by the representation layer, then apply an aggregation function to fuse the historical conversations into a fixed-length vector, and finally use a matching function to calculate a matching score \citep{DBLP:conf/sigir/YanSW16,DBLP:conf/emnlp/ZhouDWZYTLY16,DBLP:conf/aaai/XuZ021}. Interaction-based models allow historical dialogue and response candidates to interact with each other at the beginning, and they usually follow a representation-match-aggregation paradigm \citep{DBLP:conf/acl/WuWXZL17,DBLP:conf/acl/WuLCZDYZL18,DBLP:conf/coling/ZhangLZZL18}.
The PLM-based models concatenate historical dialogues and candidate options into a pre-trained multilayer self-attention network, and then perform representation, interaction, and aggregation operations in a unified manner through an attention mechanism. \citet{DBLP:conf/acl/HendersonVGCBCS19} pre-train a BERT model on a large general-domain conversation corpus, and fine-tune it in the target conversation domain, and finally aggregate each historical conversation-candidate option pair to compute a match score. \citet{DBLP:conf/cikm/GuLLLSWZ20} incorporate speaker embeddings into BERT to enable the model to perceive speaker change information. \citet{DBLP:conf/emnlp/WangZ0ZCD0L21} propose a fine-grained comparison model (FCM) that models the logical consistency between dialogue histories and generated responses. \citet{DBLP:conf/aaai/LiuFWSR021} propose a graph reasoning network (GRN) to solve the problem of insufficient reasoning ability of the model, and the performance of the model can reach close to human level. However, these models ignore the differences between options before and after reasoning and do not have sufficient reasoning ability.

In addition, there are also some models \citep{DBLP:conf/cvpr/ZhengWQZ19,DBLP:conf/emnlp/KangPLZK21} that introduce visual information into dialogue modeling, but this cannot be applied to plain text modal data.

\subsection{Graph Neural Network}
Graph neural networks (GNN) achieve excellent performance in improving the reasoning ability of the model \citep{DBLP:conf/acl/QiuXQZLZY19,DBLP:conf/acl/TuWHTHZ19}. Previous studies also apply graph convolutional networks to models to enhance the reasoning ability of the model \citep{DBLP:conf/aaai/LiuFWSR021}. Different from previous work, in order to consider the influence of different edge relations on discourse reasoning, we leverage the relational graph structure to model the sequential structure between dialogues and utilize the graph convolutional structure to enable reasoning, which has better generalization than the traditional GCN structure \citep{DBLP:conf/esws/SchlichtkrullKB18}.
\section{Model}
The architecture is shown in Figure 2, which is divided into six modules in total. Firstly, the \textbf{Context Encoder} encodes representations of historical dialogues and candidate options. Secondly, the \textbf{Option Dual Comparator (Before)} compares the representation differences of options before reasoning. Then the \textbf{Utterance Relation Reasoner (URR)} grasps the different dependencies between utterances, as well as utterances and options, and makes a reasoning between the historical dialogue and candidate options to improve the reasoning ability of the model. Next, the \textbf{Option Dual Comparator (After)} module compares the differences of the options after reasoning. Finally, the \textbf{Prediction Layer} is used to calculate the score of the options.

\subsection{Task Definition}
Given a historical dialogue $U=\{u^1,u^2,\dots,u^N\}$ where an utterance $u^n = \{w_1^n,w_2^n,\dots,w_M^n\}$ with $M$ words and a set of candidate options $O=\{o^a,o^b,o^c,o^d\}$ where $o^i$ is a candidate option. The goal is to learn the model $f(U,O)$, which can select the most logical candidate option $y$ based on the matching scores of all the candidate $O$.

\subsection{Contextual Encoder}
The context encoder mainly obtains fixed-length vector representations of options and historical conversations based on a pre-trained language model.

Given each input example $(U,O)$, the historical dialogue and all options are concatenated and fed into the pre-trained DeBERTa \citep{DBLP:conf/iclr/HeLGC21}. It is worth noting that in order to facilitate sentence-level operations of each historical dialogue and option in the following modules, we insert a [CLS] token before each utterance. Then the fixed-length representation vector for each utterance and option can be obtained, which is denoted as:
\begin{equation}\label{eq1}
	[H^U;H^O] = {\rm DeBERTa}([U;O])
\end{equation}
where $H^U \in \mathbb{R}^{|{\rm tokenize}(U)|\times d}$ and $H^O \in \mathbb{R}^{|{\rm tokenize}(O)|\times d}$ are the token-level vectors of context $U$ and options $O$, respectively. ${\rm tokenize}(\cdot)$ and $d$ are the tokenization function and hidden layer dimension of the DeBERTa model, respectively. $[;]$ represents the concatenation operation, and ${\rm DeBERTa}(\cdot)$ returns the output of the last layer of the DeBERTa model. In addition, $h_{cls}$ represents the summary vector of each utterance.

\subsection{Option Dual Comparator}
This section describes components \Rmnum{2} and \Rmnum{5} in Figure 2. The ODC aims to eliminate the interference of noisy options by comparing the differences between the options before and after reasoning based on imitating human reasoning behavior.

Two transformer \citep{DBLP:conf/nips/VaswaniSPUJGKP17} encoders are applied to serialization modeling different options before and after the URR (components \Rmnum{3} and \Rmnum{4}), so as to obtain the option representation containing the reasoning difference information, which improves the performance of the model. The multi-head attention mechanism is as follows:
\begin{equation}\label{eq2}
	\setlength{\abovedisplayskip}{3pt}
	\setlength{\belowdisplayskip}{3pt}
\begin{aligned}
	&{\rm Multihead} \\&={\rm Concat}({\rm head_1},\dots,{\rm head_h})W^O
\end{aligned}
\end{equation}
\begin{equation}\label{eq3}
\setlength{\abovedisplayskip}{3pt}
\setlength{\belowdisplayskip}{3pt}
	{\rm head_i} = {\rm Attention}(QW_i^Q,KW_i^K,VW_i^V)
\end{equation}
\begin{equation}\label{eq4}
\setlength{\abovedisplayskip}{1pt}
\setlength{\belowdisplayskip}{3pt}
	{\rm Attention}(Q,K,V)={\rm softmax}(\frac{QK^T}{\sqrt{d}})V
\end{equation}
where $W_i^Q$, $W_i^K$, $W_i^V$ and $W^O$ are all parameter matrices, and $h$ is the number of attention heads. Q, K and V are $h_{cls|o^i}$ and $h_{o^i}$, where $h_{cls|o^i}$ represents the [CLS] vector of option $o^i$, and $h_{o^i}$ represents the $o^i$ representation vector after reasoning and $i \in \{a,b,c,d\}$.
\subsection{Utterance Relational Reasoner}
\paragraph{Inter-Dependency Extraction}
Temporal and semantic dependencies between different utterances are crucial for the response selection task. Therefore we move away from explicit dependency models to implicit ones. Specifically, we believe that in the process of encoding dialogues by pre-trained language models, the temporal and semantic dependencies between different utterances are hidden in some dimensions in the semantic space, so it only needs to be "mined" and given different types.

We employ a relational attention mechanism to achieve implicit dependency modeling. 
\begin{equation}\label{eq5}
\setlength{\abovedisplayskip}{3pt}
\setlength{\belowdisplayskip}{3pt}
	q_i^s = h_{cls|s}w_i^q
\end{equation}
\begin{equation}\label{eq6}
\setlength{\abovedisplayskip}{1pt}
\setlength{\belowdisplayskip}{1pt}
	k_i^e = h_{cls|e}w_i^k
\end{equation}
\begin{equation}\label{eq7}
\setlength{\abovedisplayskip}{1pt}
\setlength{\belowdisplayskip}{1pt}
	v_i^e = h_{cls|e}w_i^v
\end{equation}
\begin{equation}\label{eq8}
\setlength{\abovedisplayskip}{1pt}
\setlength{\belowdisplayskip}{1pt}
\begin{aligned}
	&z^s_e= \frac{[q_1^s;\dots;q_n^s][k_1^e;\dots;k_n^e]^T}{\sqrt{d}}\\&*[v_1^e;\dots;v_n^e]
\end{aligned}
\end{equation}
\begin{equation}\label{eq9}
\setlength{\abovedisplayskip}{3pt}
\setlength{\belowdisplayskip}{3pt}
\begin{aligned}
	t^s_e \underset{\text{argmax}}{=} {\rm softmax}({\rm MLP}(z^s_e)))
\end{aligned}
\end{equation}
where $q_i^s$, $k_i^e$ and $v_i^e$ represent the $i$th Query of $s$, the $i$th Key and Value of $e$, respectively. $s \in [u_1,u_2,\dots,u_n]$ and $e \in [u_1,\dots,u_n,o_a,\dots,o_d]$ represent start sentence and end sentence, respectively. $w_i^q$, $w_i^k$ and $w_i^v$ are all parameter vectors. $z^s_e$ and $t^s_e \in {\rm T}$ represent the dependency vector and dependency type of $s$ to $e$, respectively. ${\rm T}$ represents the set of dependency types, and $|{\rm T}|$ represents the number of dependencies. Note that $T$ in Equation 8 represents the matrix transpose.
\paragraph{Graph Reasoning}
The goal of the reasoning module is to build a graph structure to complete the interaction between historical dialogues and candidate options. The graph structure allows messages to pass through nodes with different contextual information, which can fully consider local information for reasoning purposes. The Graph Convolutional Network (GCN) can achieve better performance in the reasoning task of QA \citep{DBLP:journals/corr/abs-1911-02170,DBLP:conf/emnlp/FangSGPWL20,DBLP:conf/acl/QiuXQZLZY19}, and the reason it works is that the GCN can summarize the feature information of the local nodes. However, in traditional GCNs, the influence of different edge relationships on nodes is not considered, which leads to the same way of aggregating neighbor node information. To avoid this, we employ relational graph convolutional networks \citep{DBLP:conf/esws/SchlichtkrullKB18}, which help the model grasp the different dependencies between utterances and between utterances and options.
The graph structure is created as follows:
\begin{compactitem}
	\item \textbf{Nodes}: The $h_{cls}$ of each utterance and option act as a node in the graph.
	\item \textbf{Edges}: There are two different ways to build edges. An edge is constructed between each historical dialogue node in the graph, and an edge is constructed between each historical dialogue node and a candidate option (Figure 2). The number of edge types is $T$, which are determined by the dependencies between the two nodes related to this edge.
\end{compactitem}

The graph modeling are now briefly described. In general, the input is a graph $G = (\mathcal{V},\xi,\mathcal{R})$ with $n$ nodes $v_i \in \mathcal{V}$, edge $e_{ij} = (v_i,v_j,r) \in \xi$, where $r \in \mathcal{R}$ is a relation type. A simple differentiable message-passing framework \citep{DBLP:conf/icml/GilmerSRVD17} is as follows:
\begin{equation}\label{eq10}
\setlength{\abovedisplayskip}{3pt}
\setlength{\belowdisplayskip}{3pt}
	\begin{aligned}
		h_i^{(l+1)}=\sigma(\sum\limits_{m\in \mathcal{M}_i}g_m(h_i^{(l)},h_j^{(l)}))	
	\end{aligned}
\end{equation}
where $h_i^{(l)} \in \mathbb{R}^{d^{(l)}}$ is the $l$th layer node representation of $v_i$ and $d^{(l)}$ is the dimensionality of $l$th layer. The $g_m(\cdot,\cdot)$ function aggregates the incoming messages and passes them through the activation function $\sigma(\cdot)$, such as the ${\rm ReLU}(\cdot)$. $\mathcal{M}_i$ is the incoming message set for node $v_i$, usually chosen as the incoming edge set. Motivated by this architecture, a multi-relational graph message propagation model is defined as:
\begin{equation}\label{eq11}
\setlength{\abovedisplayskip}{3pt}
\setlength{\belowdisplayskip}{3pt}
	\begin{aligned}
		h_i^{(l+1)}=&\sigma(\sum\limits_{r\in \mathcal{R}} \sum\limits_{j\in \mathcal{N}_i^r} \frac{1}{c_{i,r}}W_r^{(l)}h_j^{(l)}\\&+W_0^{(l)}h_i^{(l)})	
	\end{aligned}
\end{equation}
where $\mathcal{N}_i^r$ represents the set of neighbor nodes whose relationship is $r \in \mathcal{R} $ for $v_i$. $c_{i,r}$ is a problem-specific normalization constant, where $c_{i,r}=\lvert \mathcal{N}_i^r\rvert$. $W_0^{(l)}$ and $W_r^{(l)}$ are the neighbor nodes of $v_i$ and their corresponding parameter matrices respectively, which are used for linear transformation.
\subsection{Prediction Layer}
Finally, the final score is calculated by two linear layers plus an activation function, which is defined as:
\begin{equation}\label{eq12}
\setlength{\abovedisplayskip}{3pt}
\setlength{\belowdisplayskip}{3pt}
	s_{o}=W_2*{\rm ReLU}(W_1*O+b_1)+b_2	
\end{equation}
where $W_1$, $W_2$, $b_1$ and $b_2$ are trainable parameters. $O$ represents the vector of all options after ODC (After). $s_{o}$ represents the score for all options. The loss function is cross entropy loss, defined as:
\begin{equation}\label{eq13}
\setlength{\abovedisplayskip}{3pt}
\setlength{\belowdisplayskip}{3pt}
	p_i=\frac{exp(s_{o^i})}{\sum\limits_{j}exp(s_{o^i})}
\end{equation}
\begin{equation}\label{eq14}
\setlength{\abovedisplayskip}{3pt}
\setlength{\belowdisplayskip}{3pt}
	\mathcal{L} = -\sum\limits_{i}^{N}y_ilog(p_i)
\end{equation}
where $y_i$ is the true label and $N$ represents the number of samples in a batch.

\section{Experiments}
In this section, we conduct experiments on the MuTual dataset and MuTual$^{plus}$ dataset to evaluate our proposed IRRGN. In all comparative experiments, in order to ensure the authenticity of the experimental results, all training hyperparameters are kept the same. Only adjust the learning rate when the model does not converge.
\subsection{Experimental Settings}
\begin{table*}
	\centering
	\begin{tabular}{cccccccc}
		\hline
		\multirow{2}{*}{Source}& \multirow{2}{*}{Method} &\multicolumn{3}{c}{\pmb{MuTual}} &\multicolumn{3}{c}{\pmb{MuTual$^{plus}$}} \\
		\cline{3-8} 
		~&~ & \pmb{$R_4@1$} & \pmb{$R_4@2$} & \pmb{MRR} & \pmb{$R_4@1$} & \pmb{$R_4@2$} & \pmb{MRR} \\
		\hline
		\multirow{10}{1CM}{From paper \citep{DBLP:conf/acl/CuiWLZZ20}}&Random & 0.250 & 0.500 & 0.604 &0.250 &0.500 &0.604 \\
		
		~&TF-IDF \citep{DBLP:conf/sigir/Paik13} & 0.279 & 0.536 & 0.542 &0.278 &0.529 &0.764 \\
		~&Dual LSTM \citep{DBLP:conf/sigdial/LowePSP15} & 0.260 & 0.491 & 0.743 &0.251 &0.479 &0.515 \\
		~&SMN \citep{DBLP:conf/acl/WuWXZL17} & 0.299 & 0.585 & 0.595 &0.265 &0.516 &0.627 \\
		~&DAM \citep{DBLP:conf/acl/WuLCZDYZL18} & 0.241 & 0.465 & 0.518 &0.272 &0.523 &0.695 \\
		~&BERT \citep{DBLP:conf/naacl/DevlinCLT19} & 0.648 & 0.847 & 0.795 &0.514 &0.787 &0.715 \\
		~&RoBERTa \citep{DBLP:journals/corr/abs-1907-11692} & 0.713 & 0.892 & 0.836 &0.626 &0.866 &0.787 \\
		~&GPT2 \citep{radford2019language} & 0.332 & 0.602 & 0.584 &0.316 &0.574 &0.568 \\
		~&GPT2-FT \citep{radford2019language} & 0.392 & 0.670 & 0.629 &0.226 &0.611 &0.535 \\
		~&BERT-MC \citep{DBLP:conf/naacl/DevlinCLT19} & 0.667 & 0.878 & 0.810 &0.580 &0.792 &0.749 \\
		~&RoBERTa-MC \citep{DBLP:journals/corr/abs-1907-11692} & 0.686 & 0.887 & 0.822 &0.643 &0.845 &0.792 \\
		\hline
		\multirow{6}{1CM}{From Mutual leaderboard}&MUSN&0.912&0.983&0.953&\multicolumn{3}{c}{-}\\
		~&CFDR&0.913&0.986&0.954&0.735&0.904&0.849\\
		~&GRN \citep{DBLP:conf/aaai/LiuFWSR021}&0.915&0.983&0.954&0.841&0.957&0.913\\
		~&MDFN \citep{DBLP:conf/aaai/Liu0ZZ021}&0.916&\pmb{0.988}&0.956&\multicolumn{3}{c}{-}\\
		~&BIDeN&0.930&0.983&0.962&\multicolumn{3}{c}{-}\\
		~&Human & 0.938 & 0.971 & 0.964 & 0.930 & 0.972 & 0.961 \\
		\hline
		\pmb{Ours}&\pmb{IRRGN}&\pmb{0.939}&0.979&\pmb{0.965}&\pmb{0.845}&\pmb{0.962}&\pmb{0.916}\\
		\hline
	\end{tabular}
	\caption{\label{citation-guide}
		Results on the test set of the two benchmark datasets. The top half includes eleven baseline models, and the bottom half includes recent studies on these two datasets.
	}
\end{table*}
\subsubsection{Datasets}
Our proposed IRRGN is tested on MuTual and MuTual$^{plus}$ datasets\footnote{The datasets and leaderboard are available at: \small{\url{https://nealcly.github.io/MuTual-leaderboard/}}}. MuTual contains 8860 reasoning questions designed by language experts and professional annotators, which is constructed based on Chinese high school English listening test data. Each candidate is related to the historical dialogue, but only one is logically correct. MuTual$^{plus}$ is more difficult to reasoning, which uses a safe response to replace one of the four candidate options in the original dataset. MuTual$^{plus}$ is used to detect whether the model can choose a safe response when the other three candidate options are not logically correct.
\subsubsection{Metrics}
The evaluation metrics\footnote{The evaluation code is available at: \small{\url{https://github.com/Nealcly/MuTual/blob/master/eval\_sample/eval.py}}} are the same as those used in previous work. They are recall at position 1 in 4 candidate options ($R_4@1$), recall at position 2 in 4 candidate options ($R_4@2$) and Mean Reciprocal Rank (MRR) \citep{DBLP:books/aw/Baeza-YatesR99}.
\subsubsection{Baselines}
Eleven baseline models were used for comparison. Besides traditional TF-IDF \citep{DBLP:conf/sigir/Paik13} and Dual LSTM \citep{DBLP:conf/sigdial/LowePSP15}, it also includes Sequential Matching Network (SMN) \citep{DBLP:conf/acl/WuWXZL17}, Deep Attention Matching Network(DAM) \citep{DBLP:conf/acl/WuLCZDYZL18}, BERT and BERT-MC \citep{DBLP:conf/naacl/DevlinCLT19}, RoBERTa and RoBERTa-MC \citep{DBLP:journals/corr/abs-1907-11692}, GPT2 and GPT2-FT \citep{radford2019language}.
\subsubsection{Parameter Settings}
We utilize the open-source pre-trained model DeBERTa-V2$_{xxlarge}$ as the context encoder, which has 48 hidden layers, 1536 hidden-size and 24 attention heads. The $L2$ weight decays $\lambda$ is set to 0.01. The maximum sequence length is 512. We use the AdamW optimizer to optimize the model parameters with a learning rate of 2e-6. The learning rate was changed with a cosine annealing strategy in ten epochs with batch size of 2. The total number of types $T$ was set to 8. The model with the best performance on the validation set is set as the final model. We run the experiments on an A100 SXM4 GPU with 80G of memory. For more details on experimental parameter settings, please refer to our open-source code.
\subsection{Experimental Results}
\subsubsection{Comparison with baselines}
Table 1 reports the test results of IRRGN and the results of all models available for comparison. It can be observed that the $R_4@1$ metric of IRRGN significantly outperforms all compared models on both datasets, and more importantly, our proposed model outperforms human performance on all three metrics on the MuTual dataset, which proves that IRRGN has excellent reasoning ability. It is worth noting that the performance of traditional models (TF-IDF, DuLSTM, SMN and DMN) is relatively low, which indicates their insufficient reasoning ability. Pre-trained models (BERT and RoBERTa) improve in performance, but are still far behind human performance. Generative pre-trained models (GPT2) are not suitable for multi-turn dialogue reasoning problems. Our method achieves state-of-the-art performance compared to other studies (from MuTual leaderboard) on improving the reasoning ability of the model, which again validates that our method is effective. See the appendix A for the results of all baselines on the validation set.

To better verify the effectiveness of our method, we conduct ablation experiments and apply our method to other pre-trained language models for comparison.
\subsubsection{Ablation Study}
To get better insight into our IRRGN, we perform the ablation study. Specifically, five variants of IRRGN are designed: 1) \textbf{w/o ODC (After)}, the transformer encoder after the URR is removed; 2) \textbf{w/o ODC (Before)}, the transformer encoder before URR was removed; 3) \textbf{w/o ODC}, the transformer encoder before and after the URR is removed; 4) \textbf{w/o URR}, the relational attention and RGCN layers are removed. 5) \textbf{w/o ALL}, all components except pre-trained DeBERTa and Prediction are removed. The results are shown in Table 2. When ODC (After) or ODC (Before) is removed, the performance of the model decreases, which verifies the effectiveness of the dual comparison. It is worth noting that ODC (Before) appears to have a larger role than ODC (After), which is exactly what other studies overlook. When the ODC is removed, the performance of the model begins to drop significantly, which verifies that it is essential to capture the differences between the options before and after reasoning. When the URR is removed and the model shows a significant performance drop, which means that it is important for the reasoning ability of the model. Compared to using only the DeBERTa model (w/o ALL), our proposed IRRGN significantly enhances the performance on three metrics. It is worth noting that IRRGN only increases the amount of parameters by 2\%, which can be observed in the code.
\begin{table}
	\centering
	\begin{tabular}{llll}
		\hline
		Method & \pmb{$R_4@1$} & \pmb{$R_4@2$} & \pmb{MRR} \\
		
		\hline
		IRRGN & 0.931&0.972&0.959\\
		w/o ODC (After)&0.929&0.972&0.955\\
		w/o ODC (Before)&0.925&0.975&0.954\\
		w/o ODC &0.917&0.970&0.951\\
		w/o URR &0.913&0.967&0.952\\
		w/o ALL&0.904&0.964&0.946\\
		\hline
	\end{tabular}
	\caption{\label{tab2}
		Ablation experimental results of GRN on MuTual validation set. -RAO Comparison: Remove the reasoning-after option comparison module (\Rmnum{5}). -RBO Comparison: Remove the reasoning-before option comparison module (\Rmnum{2}). -ODC: Remove Option Dual Comparison module (\Rmnum{2} and \Rmnum{4}). -URR: Remove Utterance Relational Reasoner (\Rmnum{3} and \Rmnum{4}). -ALL: Remove all modules.
	}
\end{table}
\subsubsection{Generality of IRRGN}
To test the generality of the proposed IRRGN, we apply it to a widely used pre-trained language model, which includes ${\rm BERT}_{base}$, ${\rm BERT}_{large}$, ${\rm RoBERTa}_{base}$, ${\rm RoBERTa}_{large}$, ${\rm ALBERTV2}_{base}$, and ${\rm ALBERTV2}_{large}$ \citep{DBLP:conf/iclr/LanCGGSS20}. As shown in Figure 3, the performance of different pre-trained language models plus our IRRGN improves, which proves that IRRGN is generally effective.
\section{Analysis}
\subsection{Number of RGCN layers}
Table 3 shows the effect of the number of RGCN layers on the performance of the model. It can be seen that when $l=2$, the comprehensive performance of the model is the highest. This corresponds to what was analyzed in previous work \citep{DBLP:conf/iclr/KlicperaBG19}, where the number of GCN layers is related to the depth of the graph and the sparsity of the adjacency matrix. The historical dialogue turns in the MuTual and MuTual$^{plus}$ datasets are mostly within 5, which makes $l=2$ more suitable for our model.
\begin{table}
	\centering
	\begin{tabular}{cccc}
		\hline
		\makecell[c]{Number of\\ RGCN layers $l$} & \pmb{$R_4@1$} & \pmb{$R_4@2$} & \pmb{MRR} \\
		\hline
		4 & 0.900&0.975&0.945\\
		3 &0.877&0.980&0.935\\
		2&0.931&0.971&0.959\\
		1&0.882&0.972&0.935\\
		\hline
	\end{tabular}
	\caption{\label{tab3}
		Results of different number of RGCN layers on MuTual validation set.
	}
\end{table}
\begin{figure*}[t]
	\centering
	\subfigure[]{
		\includegraphics[width=0.3\linewidth]{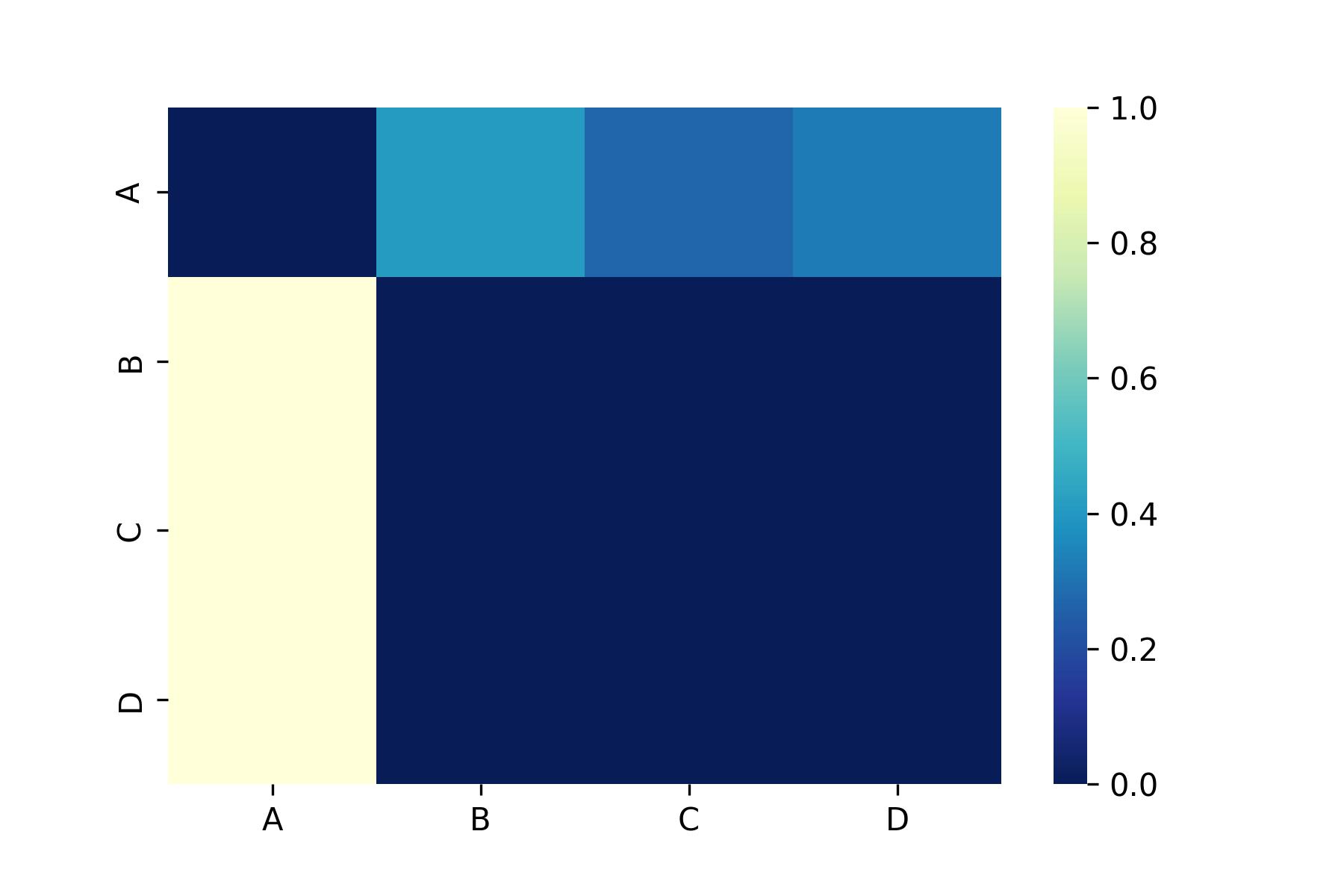}}
	\subfigure[]{
		\includegraphics[width=0.3\linewidth]{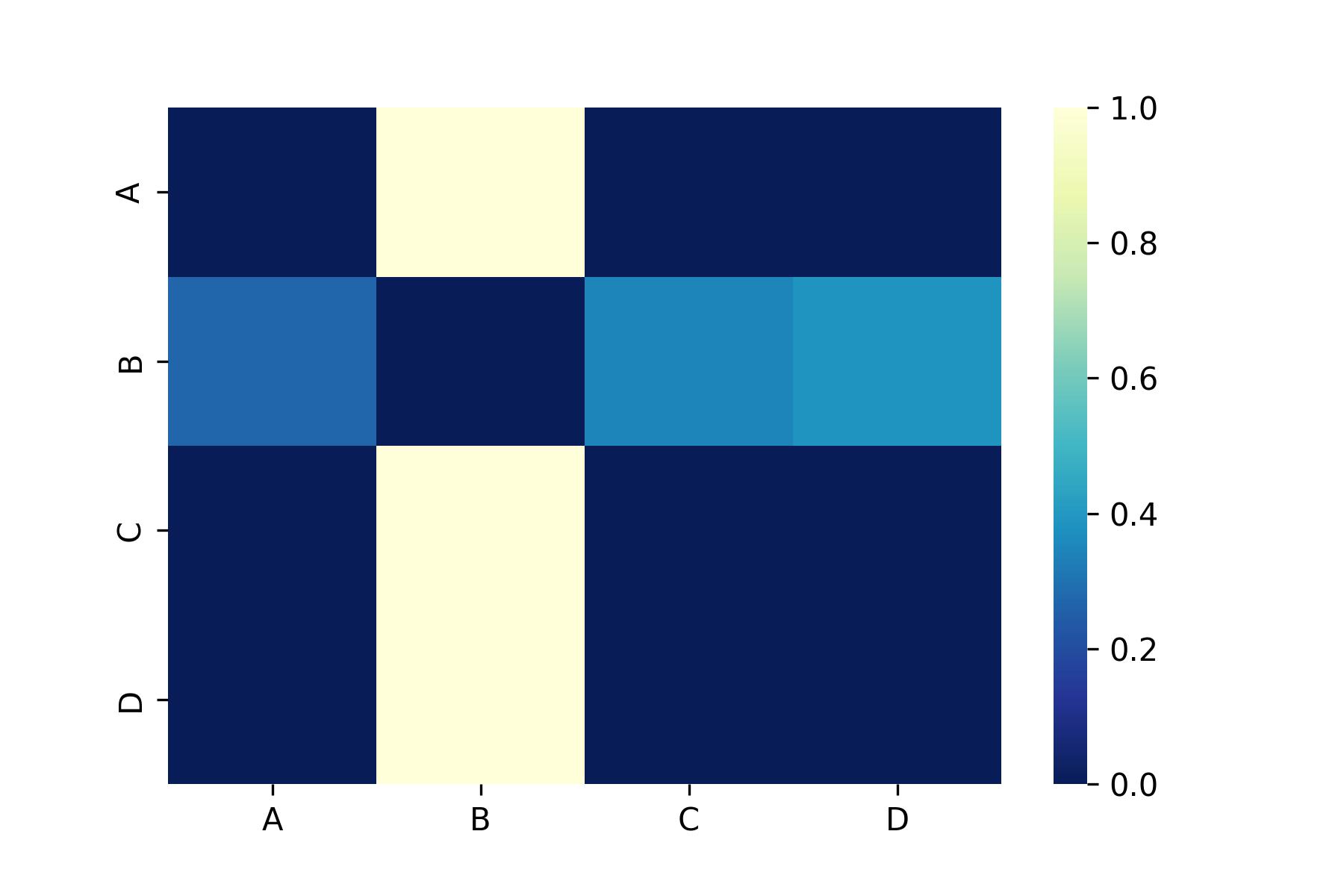}}
	\subfigure[]{
		\includegraphics[width=0.3\linewidth]{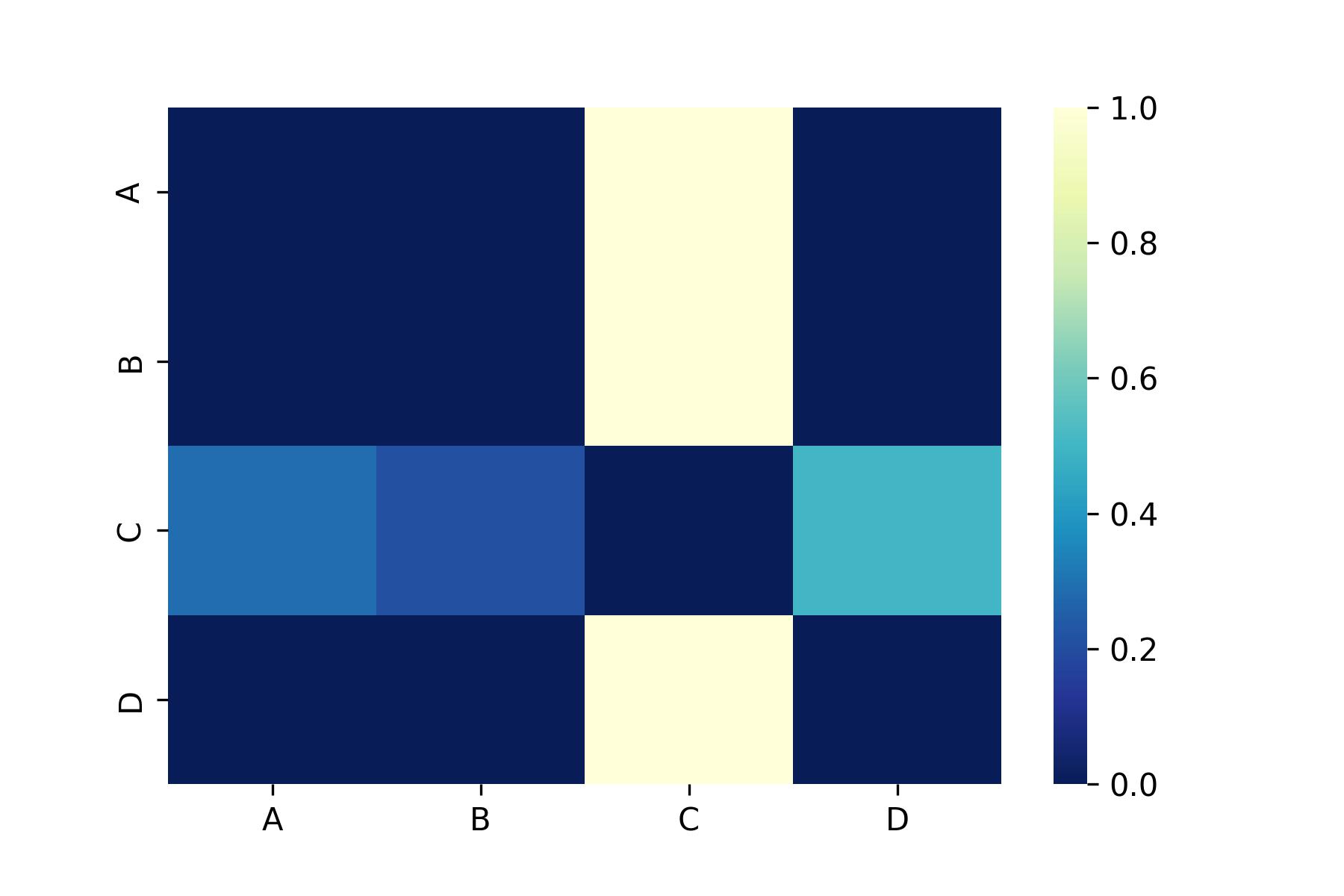}}
	\\
	\subfigure[]{
		\includegraphics[width=0.3\linewidth]{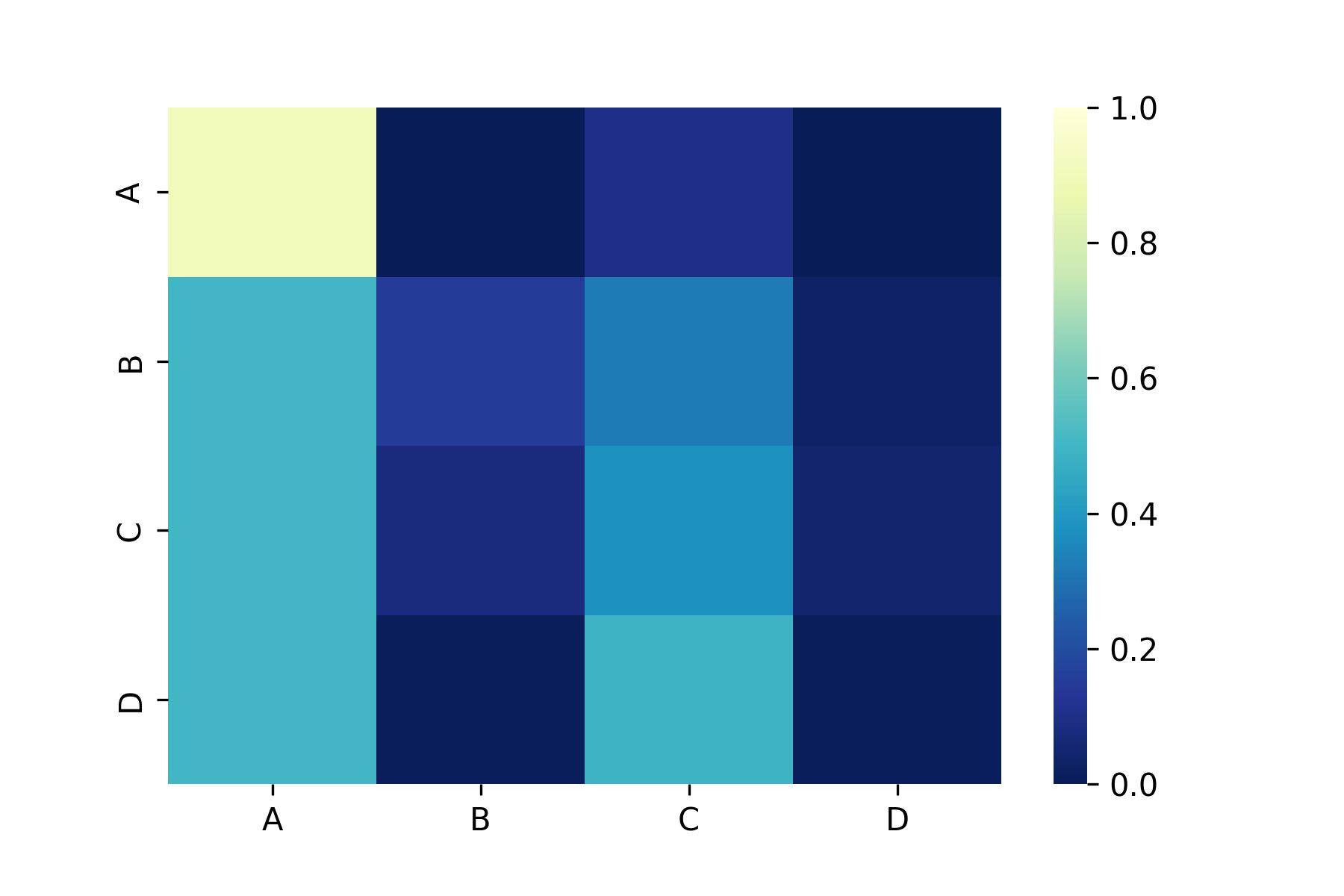}}
	\subfigure[]{
		\includegraphics[width=0.3\linewidth]{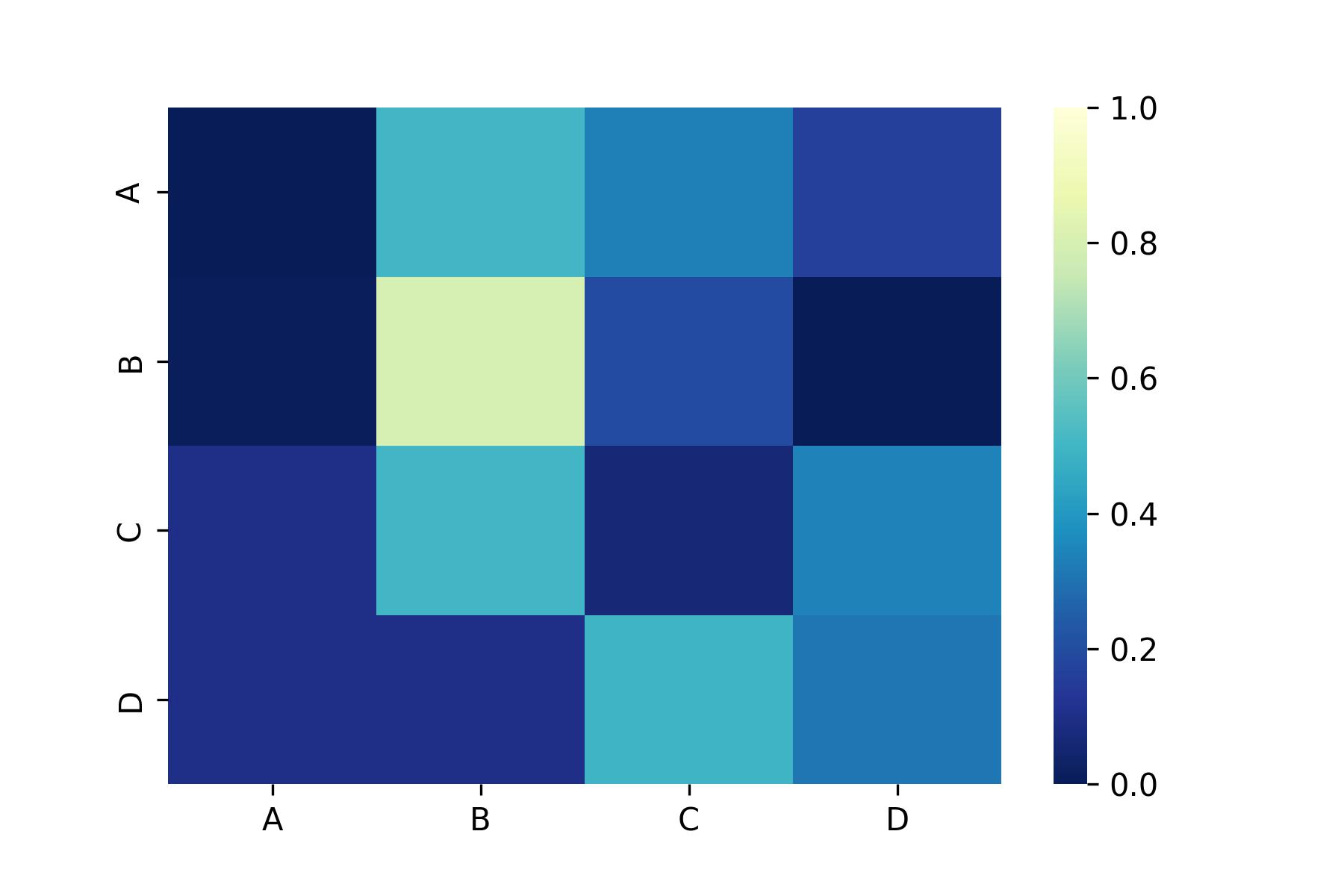}}
	\subfigure[]{
		\includegraphics[width=0.3\linewidth]{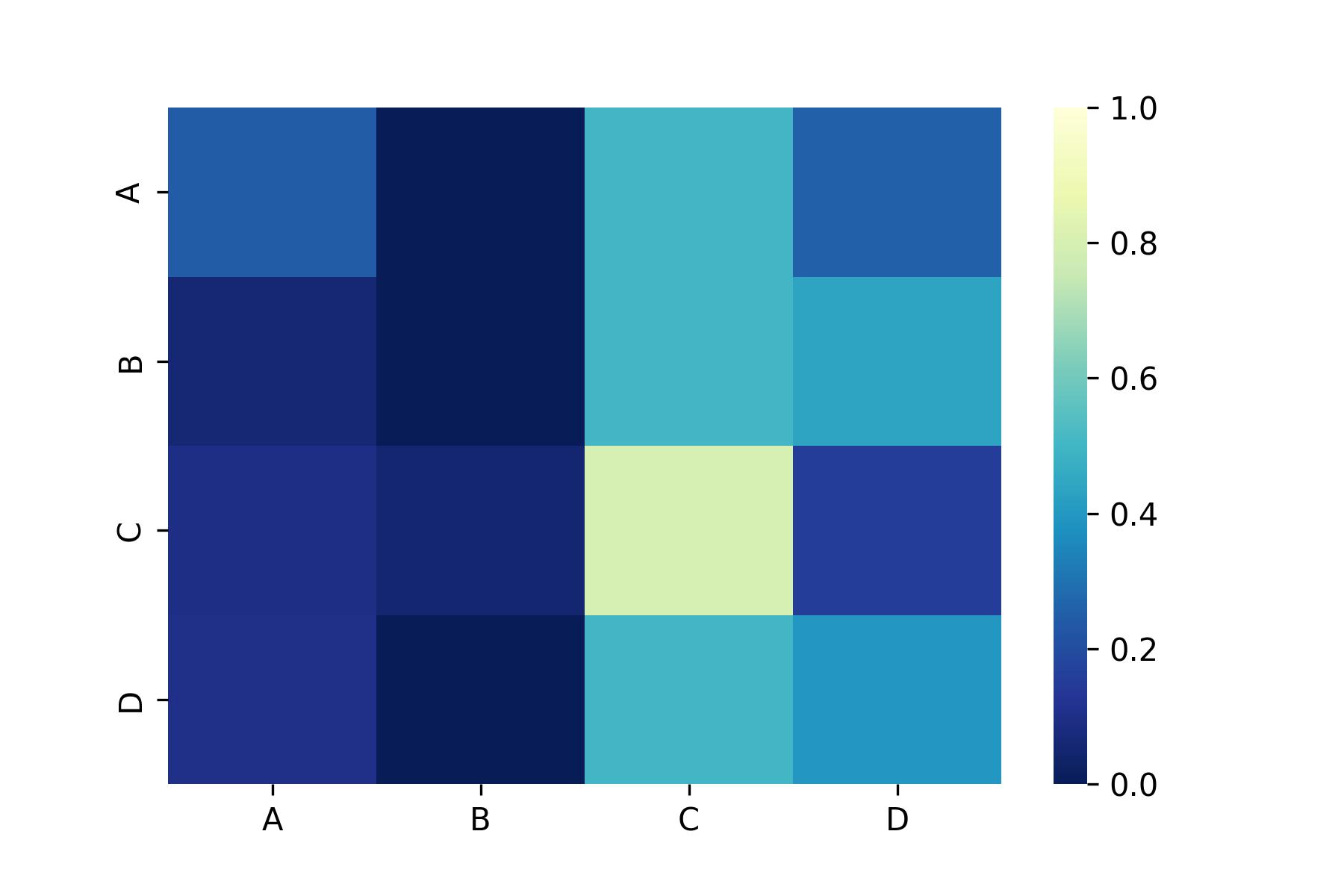}}
	\caption{Visualization of attention weights between different options. The upper row and lower row represent the attention weight maps in ODC (Before) and ODC (After), respectively. The correct answers from left to right are A, B, and C, from the example in Figure 1, dev\_1 and dev\_4, respectively.}
	\label{fig4}
\end{figure*}
\begin{figure}[t]
	\centering
	\includegraphics[scale=0.4]{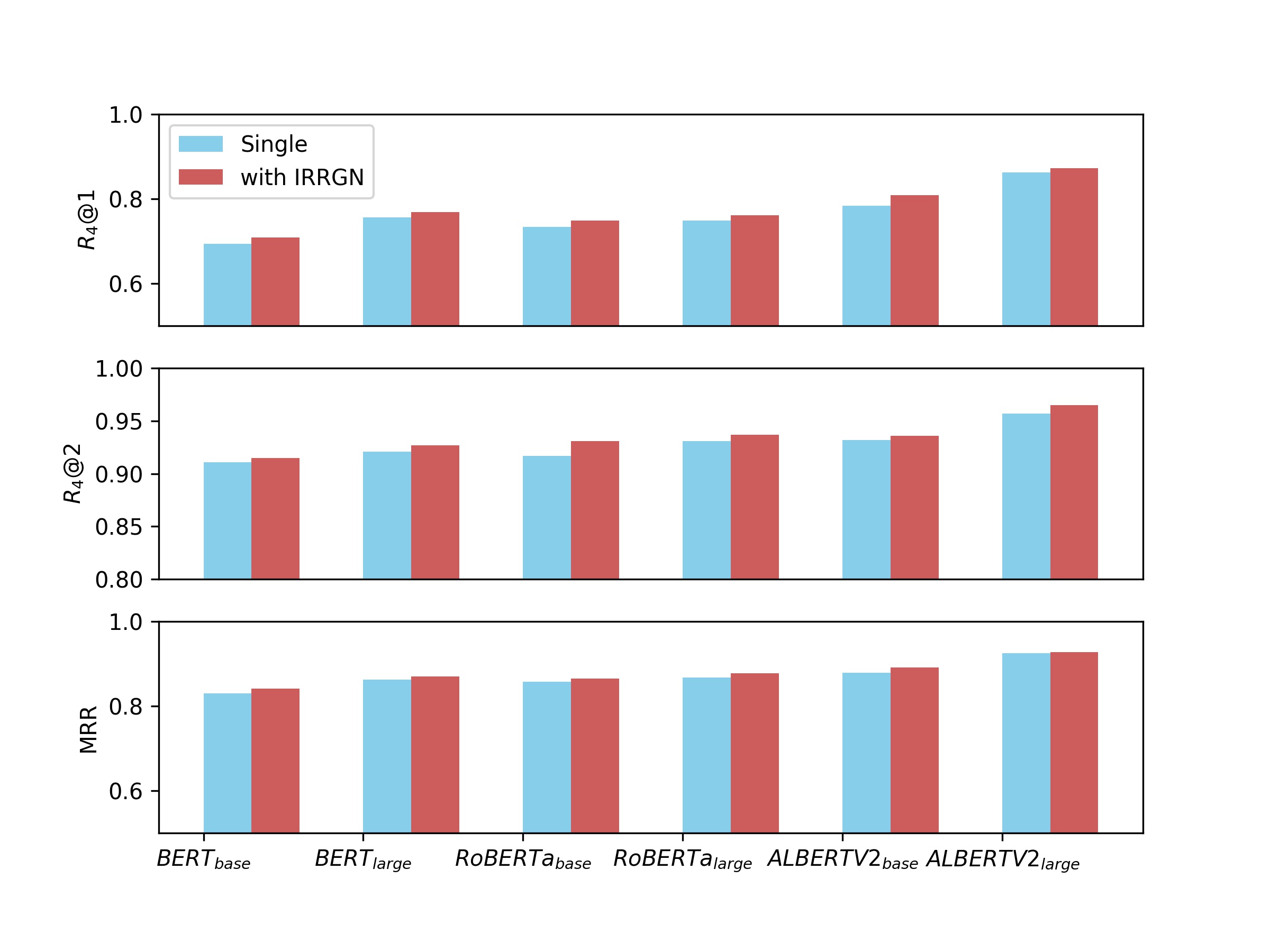}
	\caption{The $R_4@1$, $R_4@2$ and $MRR$ performance of different pre-trained language models with IRRGN on MuTual validation set.}
	\label{fig3}
\end{figure}

\subsection{Number of Arc Types}
To see the impact of the number of implicit arc types on the performance of the model, we experiment with it. As shown in Table 4, when ${\rm T} = 8$, the effect is the best. When ${\rm T} < 8$ or ${\rm T} > 8$, the performance of the model is weakened. We guess that on the MuTual and MuTual$^{plus}$ datasets, 8 different arc types can model the dependencies between utterances and between utterances and options well. When ${\rm T} < 8$, the number of arc relations is not enough to express the number of dependencies, and when ${\rm T} > 8$, too much noise are introduced, which lead to the degradation of performance of the model. When ${\rm T}=1$, the RGCN layer degenerates into the ordinary GCN layer.
\begin{table}[t]
	\centering
	\begin{tabular}{cccc}
		\hline
		\makecell[c]{Number of\\ arc types T} & \pmb{$R_4@1$} & \pmb{$R_4@2$} & \pmb{MRR} \\
		\hline
		8 & 0.931&0.971&0.959\\
		\hline
		+1 &-0.001&+0.006&-0.007\\
		+2&-0.009&+0.006&-0.004\\
		+3&-0.002&-0.007&+0.002\\
		+4&-0.011&+0.001&-0.005\\
		-1&0&+0.009&-0.002\\
		-2&-0.004&-0.002&+0.003\\
		-3&-0.002&-0.003&-0.005\\
		-4&-0.006&+0.002&+0.001\\
		\hline
	\end{tabular}
	\caption{\label{ta43}
		Results of different number of arc types on MuTual validation set. + and - represent addition and subtraction on the basis of ${\rm T}=8$ and its three metrics, respectively.
	}
\end{table}
\subsection{Effect of Option Dual Comparator}
In order to see the effect of ODC, we extract the attention weights between the options in ODC (Before) and ODC (After) for visualization, as shown in Figure 4. In the first column, correct option A in ODC (Before) focuses on options B, C, and D, which means that it is affected by the wrong options. However, in ODC (After), the correct option focuses on itself close to 1. For wrong options, they cannot focus on themselves, both in ODC (Before) and ODC (After), which shows that they can still be influenced by other options. The display of other columns is similar to the first column. It can be seen that the purpose of the ODC is to focus the correct option on itself. 

\section{Conclusion}
In this paper, we propose a novel Implicit Relational Reasoning Graph Network (IRRGN). It can implicitly define dependencies between utterances, as well as utterances and options for more efficient and flexible graph reasoning. Among other things, it captures the differences between options before and after reasoning. State-of-the-art performance is achieved on the MuTual and MuTual$^{plus}$ datasets that focus on the multi-turn dialogue reasoning task. In future work, we will further implement more fine-grained reasoning, explore model interpretability through bad cases, and let the model consider security responses.
\section{Limitations}
Although IRRGN outperforms all other models on these two datasets, there are still some points that can be improved.
\begin{compactitem}
	\item Fine-grained reasoning. Although IRRGN has excellent reasoning ability, it may not perceive more fine-grained reasoning. The nodes on the reasoning graph are at the utterance-level rather than the word-level, and we will use more Fine-grained reasoning clues to assist the dialogue selection task in the future.
	\item Security response. Like all other models, the performance of IRRGN on the MuTual$^{plus}$ dataset is lower than that of the MuTual dataset. This suggests when none of the other candidate options are logical, how to choose a security response is worth researching.
\end{compactitem}
\section*{Acknowledgements}
We thank the anoymous reviewers for valuable and inspiring comments and suggestions.
\bibliography{anthology,custom}
\bibliographystyle{acl_natbib}

\clearpage
\appendix
\section{Result of the Baselines on the Validation Set}
\label{sec:appendix}
The performance of all baselines on the validation set is shown in Table 5. The performance of the traditional model is still not high. Some models on the leaderboard achieve relatively high performance.
\begin{table*}
	\centering
	\begin{tabular}{cccccccc}
		\hline
		\multirow{2}{*}{Source}& \multirow{2}{*}{Method} &\multicolumn{3}{c}{\pmb{MuTual}} &\multicolumn{3}{c}{\pmb{MuTual$^{plus}$}} \\
		\cline{3-8} 
		~&~ & \pmb{$R_4@1$} & \pmb{$R_4@2$} & \pmb{MRR} & \pmb{$R_4@1$} & \pmb{$R_4@2$} & \pmb{MRR} \\
		\hline
		\multirow{10}{1CM}{From paper \citep{DBLP:conf/acl/CuiWLZZ20}}&Random & 0.250 & 0.500 & 0.604 &0.250 &0.500 &0.604 \\
		
		~&TF-IDF \citep{DBLP:conf/sigir/Paik13} & 0.276 & 0.541 & 0.541 &0.283 &0.530 &0.763 \\
		~&Dual LSTM \citep{DBLP:conf/sigdial/LowePSP15} & 0.266 & 0.528 & 0.538 &\multicolumn{3}{c}{-} \\
		~&SMN \citep{DBLP:conf/acl/WuWXZL17} & 0.274 & 0.524 & 0.575 &0.264 &0.524 &0.578 \\
		~&DAM \citep{DBLP:conf/acl/WuLCZDYZL18} & 0.239 & 0.463 & 0.575 &0.261 &0.520 &0.645 \\
		~&BERT \citep{DBLP:conf/naacl/DevlinCLT19} & 0.657 & 0.867 & 0.803 &0.514 &0.787 &0.715 \\
		~&RoBERTa \citep{DBLP:journals/corr/abs-1907-11692} & 0.695 & 0.878 & 0.824 &0.622 &0.853 &0.782 \\
		~&GPT2 \citep{radford2019language} & 0.335 & 0.595 & 0.586 &0.305 &0.565 &0.562 \\
		~&GPT2-FT \citep{radford2019language} & 0.398 & 0.646 & 0.628 &0.226 &0.577 &0.528 \\
		~&BERT-MC \citep{DBLP:conf/naacl/DevlinCLT19} & 0.661 & 0.871 & 0.806 &0.586 &0.791 &0.751 \\
		~&RoBERTa-MC \citep{DBLP:journals/corr/abs-1907-11692} & 0.693 & 0.887 & 0.825 &0.621 &0.830 &0.778 \\
		\hline
		\multirow{2}{1CM}{From leaderboard}
		&GRN \citep{DBLP:conf/aaai/LiuFWSR021}&0.935&0.985&0.971&\multicolumn{3}{c}{-}\\
		~&MDFN \citep{DBLP:conf/aaai/Liu0ZZ021}&0.923&0.979&0.958&\multicolumn{3}{c}{-}\\
		\hline
		\pmb{Ours}&IRRGN&0.930&0.971&0.959&0.863&0.958&0.924\\
		\hline
	\end{tabular}
	\caption{\label{tab_ap1}
		Results on the validation set of the two benchmark datasets. The top half includes eleven baseline models, and the bottom half includes recent studies on these two datasets.
	}
\end{table*}

\end{document}